\documentclass[sigconf]{acmart}
\usepackage{url}
\usepackage{enumitem}
\usepackage{placeins}
\usepackage[nolist]{acronym}
\usepackage{placeins}
\usepackage{tabularx}
\usepackage{booktabs}                   
\usepackage{multicol}                   
\usepackage{multirow}                   
\usepackage{makecell}

\AtBeginDocument{%
  \providecommand\BibTeX{{%
    \normalfont B\kern-0.5em{\scshape i\kern-0.25em b}\kern-0.8em\TeX}}}

\renewcommand\footnotetextcopyrightpermission[1]{} 
\AtBeginDocument{%
  \providecommand\BibTeX{{%
    \normalfont B\kern-0.5em{\scshape i\kern-0.25em b}\kern-0.8em\TeX}}}

\setcopyright{none}
\copyrightyear{}
\acmYear{}
\acmDOI{}
\acmISBN{}
\acmConference[HRI '24 LEAP-HRI Workshop]{HRI '24: Workshop on Lifelong Learning and Personalization in Long-Term Human-Robot Interaction (LEAP-HRI), ACM/IEEE International Conference on Human-Robot Interaction (HRI2024)}{March, 2024}{Boulder, CO.}
\acmBooktitle{HRI '24: Workshop on Lifelong Learning and Personalization in Long-Term Human-Robot Interaction (LEAP-HRI)}

\begin{document}
\title{Federated Learning of Socially Appropriate Agent Behaviours in Simulated Home Environments}

\author{Saksham Checker}
\affiliation{%
  \country{Delhi Technological University, India}
}
\email{checkersaksham01@gmail.com}

\author{Nikhil Churamani}
\affiliation{%
  \country{University of Cambridge, UK}
}
\email{nikhil.churamani@cl.cam.ac.uk}

\author{Hatice Gunes}
\affiliation{%
  \country{University of Cambridge, UK}
}
\email{hatice.gunes@cl.cam.ac.uk}





\renewcommand{\shortauthors}{Checker, Churamani and Gunes}

\begin{abstract}
As social robots become increasingly integrated into daily life, ensuring their behaviours align with social norms is crucial. For their widespread \textit{open-world} application, it is important to explore \acf{FL} settings where individual robots can learn about their unique environments while also learning from each others' experiences. In this paper, we present a novel \ac{FL} benchmark that evaluates different strategies, using multi-label regression objectives, where each client individually learns to predict the social appropriateness of different robot actions while also sharing their learning with others. Furthermore, splitting the training data by different contexts such that each client \textit{incrementally} learns across contexts, we present a novel \acf{FCL} benchmark that adapts \ac{FL}-based methods to use state-of-the-art \ac{CL} methods to continually learn socially appropriate agent behaviours under different contextual settings. \acf{FedAvg} of weights emerges as a robust \ac{FL} strategy while rehearsal-based \ac{FCL} enables \textit{incrementally learning} the social appropriateness of robot actions, across contextual splits.

\end{abstract}


\maketitle
\thispagestyle{empty}

\begin{figure*}[t!]
    \centering
    \includegraphics[width=0.92\textwidth]{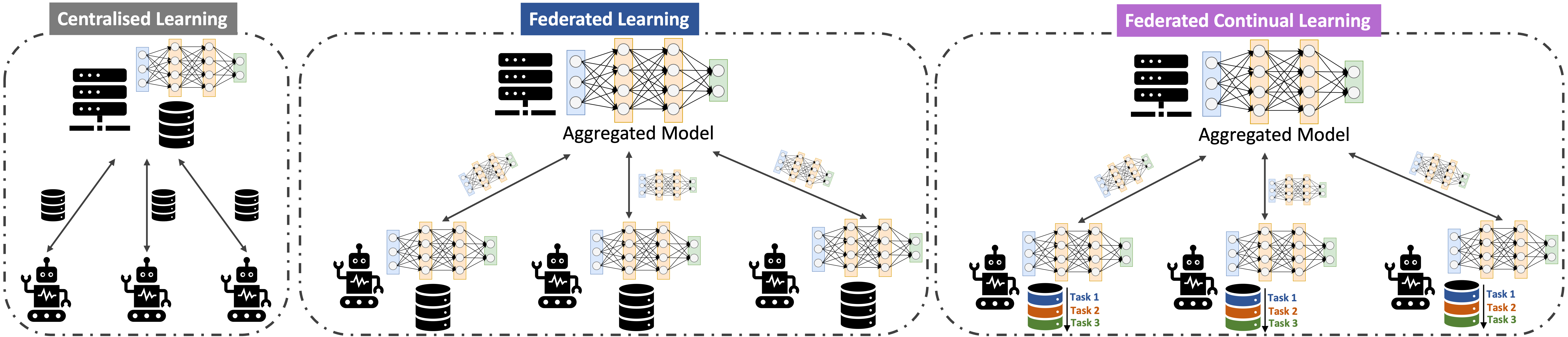}
    \caption{Centralised Learning (left) requires robots to share data with the server to train a shared model. \acf{FL} (middle) allows local model weights or gradients to be aggregated on the server without sharing data. \acf{FCL} (right) further allows individual robots to incrementally learn tasks, sharing model updates with each other.
    }
    \label{fig:learning}
\end{figure*}

\vspace{-2mm}
\section{Introduction}

Social robots deployed in \textit{open-world} human-centred environments are required to dynamically expand their knowledge by learning new tasks while preserving past knowledge~\cite{Churamani2020CL4AR, chen2018lifelong}. 
Such adaptation can enable them to support their users in day-to-day tasks by embedding themselves seamlessly within the social settings of their environments. 
Most robotic solutions for current day applications are designed as stand-alone implementations, tailored to specific tasks and/or environments~\cite{Churamani2020CL4AR}. As advances in \acf{AI} and \acf{ML} gear robots towards a more ubiquitous presence, there is a need to explore adaptive learning paradigms that to not only facilitate a widespread and generalised application but also allow individual robots to personalise towards end-user requirements and preferences. This can be in the form of several robots deployed, in a distributed manner, across different contextual settings, interacting with multiple users at a time and learning different tasks~\cite{Guerdan2023FCL}. Under such complex and diverse application settings, there is a need to move beyond centralised platforms towards more distributed learning paradigms, enabling robots to learn \textit{continually} while sharing their learning with others. 

Traditional \ac{ML}-based robotic applications, especially where multiple robots are deployed in parallel, usually follow a \textit{centralised} learning (see Figure~\ref{fig:learning};~left) approach where each robot collects data from its individual environment and communicates it to a central server. The data from each robot is then aggregated and a unified global model is trained to be used by each individual robot. Despite enabling robots to share experiences amongst each other, centralised learning approaches focus on developing a \textit{one-size-fits-all} solution by training a unified model that can generalise across applications. Data privacy becomes a major concern as each robot shares the data collected by them with the central server which may not be acceptable in certain situations. 

\acf{FL}~\cite{McMahan2017FL} (see Figure~\ref{fig:learning};~middle), on the other hand, allows robots to learn independently from their own unique experiences, updating their learning models using only the data collected by them locally. Over time, these local updates for each agent can be aggregated across the centralised server, in the form of model updates that can inform training the unified centralised model. \ac{FL} allows for a more privacy-preserving learning paradigm where local data is never shared with a centralised server. \ac{FL} solutions have been used popularly in embedded or EdgeAI devices~\cite{Imteaj2022Survey} that benefit from \textit{distributed} learning settings~\cite{Zhang2021Survey} gathering and processing their own data in their unique application settings but also sharing their learning towards training a global aggregated model that allows devices to share knowledge between each other~\cite{Li2020FedAvg}. 

As social robots interact with their environments gathering data, they need to 
efficiently discern novel knowledge from past experiences and adapt their learning models to accommodate new knowledge~\cite{Parmar2023OWL}. Under \ac{FL} settings, this means that the data collected by each robot individually need not be \acf{i.i.d}, requiring the robot to learn with sequential streams of data in an incremental manner, personalising individual robots towards their environment and users. \acf{CL}~\cite{Hadsell2020CL, Parisi2018b} can help address this problem further by enabling robots to adapt their learning with continuous and sequential streams of data acquired from non-stationary or changing environments~\cite{Churamani2020CL4AR,LESORT2020CL4R}. This may be achieved by \textit{regularising} model updates, \textit{replaying} already seen information or dynamically \textit{expanding} models to accommodate new information~\cite{Churamani2020CL4AR}. Combining \ac{FL} and \ac{CL}, \acf{FCL}~\cite{Yoon2021FCL, Guerdan2023FCL} (see Figure~\ref{fig:learning};~right) allows for individual robots, learning with sequential streams of unique local data, to also benefit from other robots' learning. Each agent periodically sends their model parameters to the centralised server where the knowledge from all agents is aggregated into a unified model which is sent back.

\begin{figure}[t]
    \centering
    \includegraphics[width=\columnwidth]{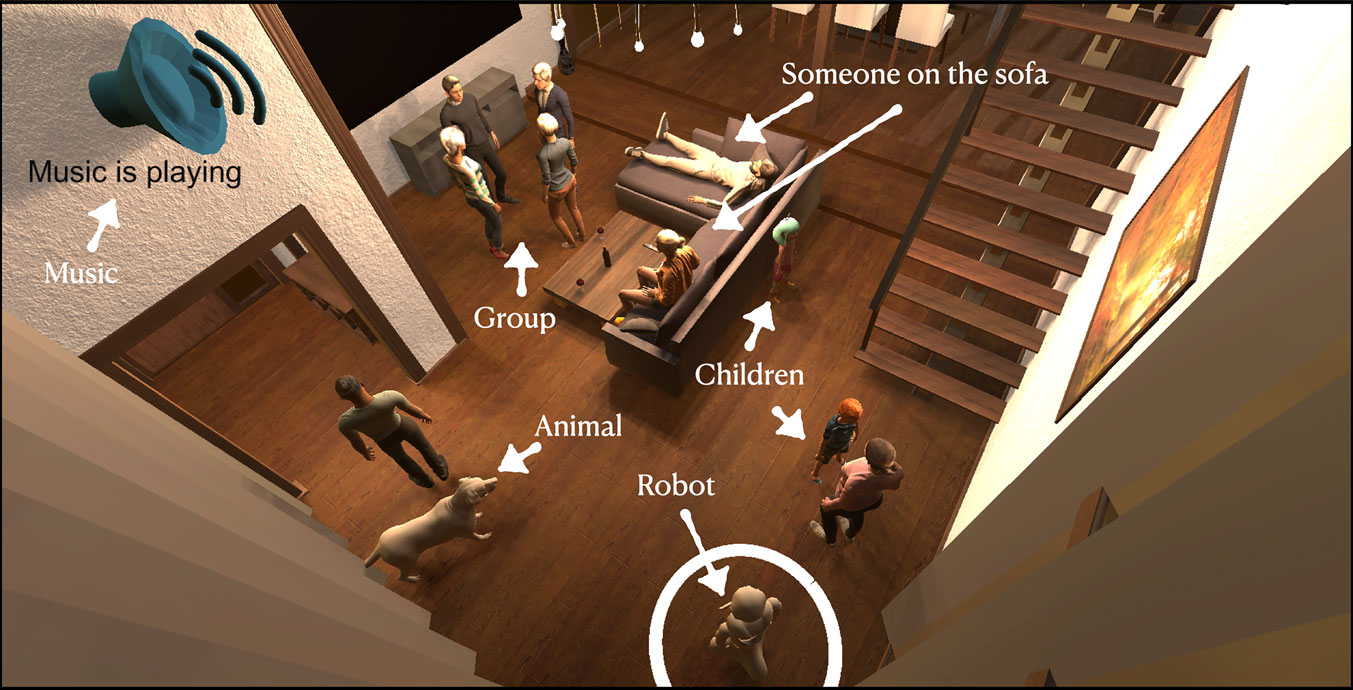}\vspace{-3mm}
    \caption{MANNERS-DB: A living room scenario with the Pepper robot. Adapted from~\cite{tjomsland2022mind}.}
    \label{fig:mannerdb}
    \vspace{-3mm}
\end{figure}

Such distributed learning settings are particularly desirable for social robots operating in human-centred environments, to understand and learn socially appropriate behaviours, depending upon the context of the interaction, environmental factors as well as individual user preferences~\cite{Churamani2020CL4AR,Guerdan2023FCL}. Whether it is effectively navigating complex social environments~\cite{Manso2020SocNav1,francis2023principles}, learning approach and positioning behaviours~\cite{gao2019learning, McQuillin2022RoboWaiter} or learning task-specific behaviours~\cite{tjomsland2022mind}, it is essential for robots to consider the social-appropriateness of their behaviours in order to comply with social norms~\cite{Ayub2020What, cramer2010effects}. 

In this paper, we explore simulated environments with humans and robotic agents to learn the social appropriateness of different high-level tasks as a use case for \ac{FL} and \ac{FCL}-based application of open-world learning. Depending upon the user, context or social norms~\cite{Ayub2020What}, the agents need to learn what actions may be appropriate for them to perform and how they will be viewed by their users~\cite{tjomsland2022mind, McQuillin2022RoboWaiter}. Here, we explore the MANNERS-DB dataset~\cite{tjomsland2022mind} that provides social-appropriateness ratings for different agent actions in simulated home settings. We benchmark different \ac{FL} and \ac{FCL} methods to understand how such a learning of socially-appropriate agent behaviours can be realised in a distributed manner (\ac{FL}) learning incrementally and sequentially (\ac{FCL}), where individual agents effectively share their learning with each other.

\section{Methodology}

\subsection{Learning Scenario: MANNERS-DB Dataset}
Learning socially appropriate behaviours in complex home settings requires robots to be sensitive to its positioning with respect to other objects and users as well as individual user preferences. In this work, we explore a simulated living room scenario consisting of different actors where the agent is tasked upon learning the social appropriateness of different actions. For this, the MANNERS-DB dataset~\cite{tjomsland2022mind} is used that consists of  $3$D scenes, created with Unity, of the Pepper robot co-inhabiting a living room space with other humans (adults and children) and animals under different social settings (see Figure~\ref{fig:mannerdb}). For each scene, the robot can perform $8$ different tasks, that is, \textit{vacuuming, mopping, carrying warm/cold food, carrying big/small objects, carrying drinks or cleaning/starting conversations}. These tasks can be performed by the robot either within a circle of influence or in the direction of operation with the only difference being cleaning within a circle and starting a conversation in the direction of the arrow. Crowd-sourced annotations are provided for the social appropriateness of each of these actions for every scene (out of a total of $\approx 1000$ scenes), labelled on a $5$-point Likert scale, ranging from very inappropriate to very appropriate.

\vspace{-2mm}
\subsection{Experimentation Settings}
\subsubsection{Input Features and Data Augmentation:} 
For each scene, a $29-$dimensional descriptor is provided consisting of \textit{features} such as a flags for circle of influence or direction of operation, number of humans, children and animals, distance between the robot and the 3 closest humans, amongst others (see~\cite{tjomsland2022mind} for the complete list). We use the $29-$d scene descriptors as the input to the model to predict the social-appropriateness of each of the $8$ actions (within the circle or in the direction of the arrow). For both \ac{FL} and \ac{FCL} evaluations, the data is split into training and test splits in the ratio of $75\%:25\%$. The training data is further split amongst the different clients ($2$ or $10$) with the shared test-set used for evaluation. For \ac{FCL} evaluations, the training set of individual clients (or simulation nodes) is further split into two tasks, that is, samples depicting the robot operating with an circle (Task~$1$) and in the direction of the arrow (Task~$2$). Since the MANNERS-DB dataset is a relatively small dataset with approx. $1000$ samples, we also benchmark the different \ac{FL} and \ac{FCL} methods using data augmentation as well. For this, a Gaussian noise ($\mu=0, \sigma=0.01$) is added to each feature.

\vspace{-2mm}
\subsubsection{Implementation Details:} 
For each \ac{FL} and \ac{FCL} approach, a \acf{MLP}-based model is implemented consisting of two \ac{FC} layers of $16$ units each with a \textit{linear} activation. Each \ac{FC} layer is followed by a \verb|BatchNormalisation| layer. The output of the last \ac{FC} layer is passed to the $8-$unit output layer, predicting the social appropriateness for each of the $8$ robot actions. The experiments are run for $2-10$ clients. This relatively low number of clients is to compare these methods for a potential real-world evaluation to be conducted using physical robots. For brevity, results for only $2$ and $10$ clients are presented. 
All models are implemented using the PyTorch \footnote{https://pytorch.org} and Flower\footnote{https://flower.dev} Python Libraries. 

\vspace{-1mm}

\subsubsection{Evaluation Metrics:}
Since each client is learning the social-appropriateness of each of the $8$ possible actions, we use regression-based loss and evaluation metrics. The models are trained using the \ac{MSE}, computed as an average across the $8$ actions. Furthermore, the average \ac{RMSE} and \ac{PCC}~\cite{benesty2009pearson} scores are also reported, computed as an average across the $8$ actions.

\section{The FL Benchmark}
We compare different state-of-the-art \ac{FL} methods, both without and with data augmentation, presenting a novel benchmark for learning social appropriateness for different robot actions simultaneously.

\vspace{-2mm}
\subsection{Compared approaches}

\begin{table}[t!]
    \caption{Federated Learning results for the MANNERS-DB dataset for two (left) and ten (right) clients. Bold values denote best while \textit{[bracketed]} denote second-best values.}
    \label{tab:FL}
    \setlength{\tabcolsep}{4.0pt}

    \small
    \centering
    \vspace{-3mm}

    \begin{tabular}{l|rrr|rrr}\toprule
        & \multicolumn{3}{c|}{\textit{Two Clients}} &   \multicolumn{3}{c}{\textit{Ten Clients}}\\\cline{2-7}

        Method          & Loss              & RMSE              & PCC               & Loss              & RMSE              & PCC               \\\hline

        & \multicolumn{6}{c}{\textit{W/O Augmentation}}\\\hline
        FedAvg          & \textbf{0.219}    & \textbf{0.468}    & 0.445             & [0.225]             & 0.475             & 0.453             \\
        FedBN           & 0.222             & 0.471             & [0.456]             & 0.226             & 0.475             & [0.454]             \\
        FedProx         & [0.220]             & [0.469]             & \textbf{0.458}    & [0.225]             & [0.474]             &\textbf{ 0.465}    \\
        FedOpt          & 0.223             & 0.472             & 0.448             & \textbf{0.224}    & \textbf{0.473}    & 0.452             \\
        FedDistill      & 0.250             & 0.500             & 0.425             & 0.670             & 0.818             & 0.419             \\\hline
        & \multicolumn{6}{c}{\textit{W/ Augmentation}}\\\hline
        FedAvg$_{Aug}$      & \textbf{0.212}             & \textbf{0.460}             & \textbf{0.421}             & \textbf{0.213}             & \textbf{0.462}             & \textbf{0.419}             \\
        FedBN$_{Aug}$      & 0.231             & 0.480             & [0.402]             & [0.224]             & [0.473]             & 0.402             \\
        FedProx$_{Aug}$     & [0.222]             & [0.471]             & 0.401             & 0.227             & 0.476             & [0.404]             \\
        FedOpt$_{Aug}$      & 0.231             & 0.481             & 0.397             & 0.227             & 0.477             & 0.401             \\
        FedDistill$_{Aug}$  & 0.251             & 0.501             & 0.383             & 0.231             & 0.481             & 0.399             \\\bottomrule
    \end{tabular}
    \vspace{-3mm}
\end{table}

\begin{table*}[t!]
    \caption{Federated Continual Learning results for the MANNERS-DB dataset for two (left) and ten (right) clients. Data is split into two tasks: Circle (Task 1) and Arrow (Task 2). Bold values denote best while \textit{[bracketed]} denote second-best values.}
    \label{tab:FCL}
    \centering
    \small

    \vspace{-2.5mm}

    \begin{tabular}{l|rrr|rrr|rrr|rrr}\toprule
        & \multicolumn{6}{c|}{\textit{Two Clients}} &   \multicolumn{6}{c}{\textit{Ten Clients}}\\\cline{2-13}

        & \multicolumn{3}{c|}{\textit{After Task 1}} &   \multicolumn{3}{c|}{\textit{After Task 2}} & \multicolumn{3}{c|}{\textit{After Task 1}} &   \multicolumn{3}{c}{\textit{After Task 2}}\\\cline{2-13}

        Method          & Loss              & RMSE              & PCC               & Loss              & RMSE              & PCC               & Loss              & RMSE              & PCC               & Loss              & RMSE              & PCC\\\hline

        & \multicolumn{12}{c}{\textit{W/O Augmentation}}\\\hline
        
        FedAvg$_{EWC}$      & [0.248]             & [0.492]             & 0.562             & 0.265             & 0.503             & 0.579            & 0.263 & 0.504 & 0.543 & 0.262 & 0.502 & 0.557 \\
        
        FedAvg$_{EWCOnline}$ & \textbf{0.242} & \textbf{0.488} &\textbf{ 0.621} & 0.261 & 0.501 & 0.581 & \textbf{0.249} & \textbf{0.491} & \textbf{0.582} & 0.268 & 0.507 & 0.550 \\
                
        FedAvg$_{SI}$ & 0.258 & 0.502 & [0.577] & [0.240] & 0.483 & [0.585] & 0.262 & 0.505 & 0.546 & \textbf{0.249} & \textbf{0.489} & [0.574] \\

        FedAvg$_{MAS}$ & 0.271 & 0.513 & 0.567 & 0.241 & [0.481] & \textbf{0.589 } & [0.251] & [0.494] & 0.573 & 0.261 & 0.502 & 0.556 \\

        FedAvg$_{NR}$ & 0.262 & 0.503 & 0.531 &\textbf{ 0.235} & \textbf{0.480} & 0.564 & 0.252 & [0.494] & [0.577] & [0.250] & [0.494] & \textbf{0.583} \\\hline

        &\multicolumn{12}{c}{\textit{W/ Data-augmentation}}\\\hline
        
        FedAvg$_{EWC}$ & 0.197 & 0.439 & \textbf{0.631} & 0.240 & 0.482 & 0.528 & \textbf{0.179} & \textbf{0.419} & 0.620 & [0.226] & [0.470] & [0.541] \\
        
        FedAvg$_{EWCOnline}$ & \textbf{0.188 }& \textbf{0.429} & 0.619 & 0.251 & 0.491 & 0.530 & [0.180] & [0.420] & 0.634 & [0.226] & [0.470] & \textbf{0.542} \\
              
        FedAvg$_{SI}$   & 0.195 & 0.438 & [0.622] & 0.236 & 0.481 & [0.538] & 0.184 & 0.425 & [0.637] & 0.240 & 0.481 & 0.529 \\

        FedAvg$_{MAS}$ & [0.192] & [0.434] & 0.621 & [0.232] & [0.475] & 0.534 & 0.184 & 0.424 & \textbf{0.640} & 0.232 & 0.474 & 0.534 \\

        FedAvg$_{NR}$ & 0.205 & 0.447 & 0.613 &\textbf{ 0.211} & \textbf{0.456} & \textbf{0.550} & 0.182 & 0.422 & 0.631 & \textbf{0.221} & \textbf{0.466} & 0.540 \\\bottomrule
    \end{tabular}
    \vspace{-3mm}
\end{table*}

\subsubsection{\textbf{FedAvg:}} \acf{FedAvg}~\cite{Li2020FedAvg} is a straightforward approach for weight aggregation across clients in 
\textit{rounds} where at each round, a centralised server gathers the model weights from $m$ clients, aggregates them by computing the average across these $m$ clients to form the \textit{global model weights} and then updates the weights of each client with the global model weights.
\vspace{-2mm}

\subsubsection{\textbf{\acs{FedBN}:}} One of the main problems with \ac{FedAvg} comes under \textit{heterogeneous} data conditions where local data is non-\ac{i.i.d}. FedBN~\cite{Li2021FedBN} aims to address this problem by adapting \ac{FedAvg} by keeping the parameters for all the \verb|BatchNormalisation| layers `strictly local', that is, all other model weights are aggregated across clients apart from the \verb|BatchNormalisation| parameters. 
\vspace{-2mm}

\subsubsection{\textbf{FedProx:}} Similar to FedBN, FedProx~\cite{Li2020FedProx} also proposes improvements over \ac{FedAvg} by allowing for only partial aggregation of weights by adding a proximal term to \ac{FedAvg}. The objective for each client is thus modified to minimise $F_k (\omega) + \frac{\mu}{2}||\omega-\omega^t||^2$ where $F_k$ is the loss, $\omega$ are the local model weights to optimise and $\omega^t$ are the global parameters at time-step $t$. \ac{FedAvg} can be considered to be a special case of FedProx with $\mu=0$.
\vspace{-2mm}

\subsubsection{\textbf{FedOpt:}} Another challenge faced by \ac{FedAvg} is that of \textit{adaptivity}.  To address this, FedOpt~\cite{Reddi2021FedOpt} is proposed as a `general optimisation framework' where each client uses a \textit{client optimiser} to optimise on local data while the server updates apply a gradient-based \textit{server optimiser} to the aggregated model weights. We use the Adam optimiser for both client and server optimisation. \ac{FedAvg} can be considered to be a special case of FedOpt where both client and server optimisers use \verb|StochasticGradientDescent| (SGD) with server learning rate set to $1$. 
\vspace{-2mm}

\subsubsection{\textbf{FedDistill:}} The FedDistill~\cite{Jiang2020FedDistill} approach also aims to improve the ability of the clients to deal with \textit{heterogeneous} data conditions by using \textit{knowledge distillation}~\cite{Hinton2015Distillation}. Each client maintains two models: (i) a local copy of the global model and (ii) a personalised model that acts as a teacher to the student global model. The updated student model is then aggregated across clients. 


\noindent For the above-mentioned approaches, in our experiments, each client undergoes $10$ aggregation rounds and test-metrics are computed at the end of each round using the aggregated global model. 

\vspace{-1mm}
\subsection{Results and Discussion}
Table~\ref{tab:FL} presents the \ac{FL} benchmark results. For $2$ clients, the train-set is split into two equal parts while for $10$ clients, it is split into ten equal parts. Thus, when evaluating models without augmentation, there is relatively more data per client for two clients compared to ten clients. We see that \ac{FedAvg} performs the best under such settings, especially as each sample enabling learning across all 8 actions, with FedProx being a close second. For $10$ clients however, adaptive optimisers under FedOpt are able to work well with low amount of per-client data with FedProx performing the second best. Similar trends are witnessed in evaluations with augmentation where a relatively large amount of data is available for all clients under the $2$ and $10$-client splits. \ac{FedAvg} performs the best here as well while FedProx and FedBN are the next best approaches. Our evaluations presents a multi-label regression problem which is different from classification where \ac{FedAvg} offers a relatively simple and robust learning methodology to predict the social appropriateness of agent behaviours in the MANNERS-DB dataset. Even in situations when FedAvg is not the best performing approach the difference in model performances is marginal.

\section{The FCL Benchmark}
As can be seen in the \ac{FL} results (see Table~\ref{tab:FL}), \ac{FedAvg} emerges as a simple and robust approach in our multi-label regression set-up. For incremental learning across tasks under non-\ac{i.i.d} settings, we adapt \ac{FedAvg} for \ac{FCL} using different state-of-the-art \ac{CL}-based objectives proposing \ac{FCL} variants for \ac{FedAvg}. Each client incrementally learns the social appropriateness for different robot actions under different learning contexts using the following \ac{CL} methods, followed by a weight aggregation round where model weights are averaged across clients. We focus primarily on regularisation-based \ac{CL} as these methods do not require additional computational resources. \ac{NR}~\cite{Hsu18_EvalCLNR} is included as a baseline for rehearsal-based methods. All methods are compared, both without and with data augmentation, presenting a novel \ac{FCL} benchmark. 

\vspace{-2mm}
\subsection{Compared Approaches}

\subsubsection{\textbf{FedAvg$_{EWC}$:}} The \acf{EWC}~\cite{kirkpatrick2017overcomingEWC} approach introduces quadratic penalties on weight updates between old and new tasks. For each parameter, an importance value is computed using that task's training data, approximated as a \textit{Gaussian Distribution} with its mean as the task parameters and the importance determined by the diagonal of the Fischer Information Matrix. 

\vspace{-2mm}

\subsubsection{\textbf{FedAvg$_{EWCOnline}$:}} \ac{EWC}Online~\cite{schwarz2018progressEWCOnline} offers an improvement over \ac{EWC} where, instead of maintaining individual quadratic penalty terms for each of the tasks, a \textit{running sum} of the Fischer Information Matrices for the previous tasks is maintained. 
\vspace{-2mm}

\subsubsection{\textbf{FedAvg$_{SI}$:}} The \ac{SI}~\cite{zenke2017continualSI} approach penalises changes to weight parameters or synapses such that new tasks can be learnt without forgetting the old. To avoid forgetting, importance for solving a learned task is computed for each parameter and changes in important parameters are discouraged.
\vspace{-2mm}

\subsubsection{\textbf{FedAvg$_{MAS}$:}} The \acf{MAS}~\cite{aljundi2018memoryMAS} approach attempts to alleviate forgetting by calculating an importance value for each parameter by examining the sensitivity of the output function instead of the loss function. Higher the impact of changes to a parameter, higher is the importance assigned and higher is the penalty imposed. Yet, different from \ac{EWC} and \ac{SI}, parameter importance is calculated using only unlabelled data.
\vspace{-2mm}

\subsubsection{\textbf{FedAvg$_{NR}$:}} For the \acf{NR}~\cite{Hsu18_EvalCLNR} approach, each client maintains a replay buffer where a fraction of previously seen data is stored. This \textit{old data} is interleaved with the new data to create mixed mini-batches to train the model by simulating \ac{i.i.d} data settings in an attempt to mitigate forgetting in the model.

\noindent The above-mentioned \ac{CL}-based adaptations to the FedAvg approach are applied locally for each client. The importance values for \ac{EWC}, \ac{EWC}Online, \ac{SI} and \ac{MAS} approaches are calculated before 
aggregation across clients. The computed importance values for each of the parameters are then used to penalise changes in local weight updates between tasks, mitigating \textit{forgetting}. 

\vspace{-1mm}
\subsection{Results and Discussion}
Table~\ref{tab:FCL} presents the \ac{FCL} results. For each approach, average test-set metrics are calculated using the aggregated global model after all aggregation rounds are completed for each task. Task~$1$ results represent test-set results only for data pertaining to the `circle' split while the entire test-set is used to evaluate the models after Task~$2$. Learning incrementally is seen to have a positive effect on model performance. This is evidenced from the average \ac{PCC} values (\textit{after task 2}) being better for \ac{FCL} vs. \ac{FL} evaluations, for both $2$ and $10$ clients. Without using data augmentation, rehearsal-based \ac{NR} approach performs better than regularisation-based approaches after witnessing both tasks as it maintains a memory buffer to store previously seen task $1$ samples. With the relatively low number of samples in the MANNERS-DB dataset, almost all the samples from task~$1$ can be maintained in the memory buffer, resulting in the better performance scores for \ac{FedAvg}$_{NR}$. For $10$ clients, \ac{SI} comes closer however \ac{NR} still achieves the best \ac{PCC} scores. A similar trend is seen with data augmentation as well where \ac{NR} still is able to retain past knowledge the best. Yet, as a separate memory buffer needs to be maintained for \ac{NR}, it may not be the most resource-efficient approach. This may become particularly challenging when dealing with high-dimensional data such as images or videos~\cite{Stoychev2023LGR}.

\section{Conclusions and Future Work}
This work presents a novel benchmark for learning socially appropriate robot behaviours in home settings comparing different \ac{FL} and \ac{FCL} approaches. For \ac{FL} evaluations, \ac{FedAvg} offers a relatively simple and robust learning methodology matching baseline evaluation scores from traditional \ac{ML}-based methods~\cite{tjomsland2022mind}. This motivates the use of \ac{FedAvg} to be adapted for \ac{FCL} evaluations when incrementally learning different tasks. Our \ac{FCL} evaluations show that rehearsal-based \ac{NR} approach is best suited for such applications albeit being memory intensive. In this work, we explore pre-extracted $29-$d scene descriptions to predict the social appropriateness of different robot actions. Our future work will focus on end-to-end learning directly using scene renders while exploring more resource-efficient generative feature replay methods~\cite{Stoychev2023LGR, Liu2020Geneative}.

\begin{acks}

\textbf{Funding:} S.~Checker contributed to this work while undertaking a remote visiting studentship at the Department of Computer Science and Technology, University of Cambridge. N.~Churamani and H.~Gunes are supported by Google under the GIG Funding Scheme. 

\noindent \textbf{Open Access:} For open access purposes, the authors have applied a Creative Commons Attribution (CC BY) licence to any Author Accepted Manuscript version arising.

\noindent \textbf{Data Access Statement:} This study involves secondary analyses of the existing datasets, that are described and cited in the text. 

\noindent \textbf{Code Access:} \url{https://github.com/nchuramani/FCL-MannersDB}.
\end{acks}

\begin{acronym}
    \acro{AI}{Artificial Intelligence}

    \acro{CF}{Catastrophic Forgetting}
    \acro{CL}{Continual Learning}
    \acro{Class-IL}{Class-Incremental Learning}
    
    \acro{EWC}{Elastic Weight Consolidation}

    \acro{FL}{Federated Learning}
    \acro{FC}{Fully Connected}
    \acro{FCL}{Federated Continual Learning}
    \acro{FedAvg}{Federated Averaging}
    \acro{FedBN}{Federated BatchNorm}
    \acro{HRI}{Human-Robot Interaction}

    \acro{i.i.d}{\textit{independent and identically distributed}}

    \acro{LGR}{Latent Generative Replay}

    \acro{MAS}{Memory Aware Synapses}
    \acro{ML}{Machine Learning}
    \acro{MLP}{Multilayer Perceptron}
    \acro{MSE}{Mean Squared Error}
    
    \acro{NC}{New Concepts}
    \acro{NI}{New Instances}
    \acro{NIC}{New Instances and Concepts}
    \acro{NR}{Naive Rehearsal}

    \acro{PCC}{Pearson's Correlation Coefficient}
    
    \acro{RL}{Reinforcement Learning}
    \acro{RMSE}{Root Mean Squared Error}

    \acro{SI}{Synaptic Intelligence}
    
    \acro{Task-IL}{Task-Incremental Learning}

\end{acronym}
\bibliographystyle{ACM-Reference-Format}
\bibliography{main}


\begin{thebibliography}{32}


\ifx \showCODEN    \undefined \def \showCODEN     #1{\unskip}     \fi
\ifx \showDOI      \undefined \def \showDOI       #1{#1}\fi
\ifx \showISBNx    \undefined \def \showISBNx     #1{\unskip}     \fi
\ifx \showISBNxiii \undefined \def \showISBNxiii  #1{\unskip}     \fi
\ifx \showISSN     \undefined \def \showISSN      #1{\unskip}     \fi
\ifx \showLCCN     \undefined \def \showLCCN      #1{\unskip}     \fi
\ifx \shownote     \undefined \def \shownote      #1{#1}          \fi
\ifx \showarticletitle \undefined \def \showarticletitle #1{#1}   \fi
\ifx \showURL      \undefined \def \showURL       {\relax}        \fi
\providecommand\bibfield[2]{#2}
\providecommand\bibinfo[2]{#2}
\providecommand\natexlab[1]{#1}
\providecommand\showeprint[2][]{arXiv:#2}

\bibitem[Aljundi et~al\mbox{.}(2018)]%
        {aljundi2018memoryMAS}
\bibfield{author}{\bibinfo{person}{Rahaf Aljundi}, \bibinfo{person}{Francesca
  Babiloni}, \bibinfo{person}{Mohamed Elhoseiny}, \bibinfo{person}{Marcus
  Rohrbach}, {and} \bibinfo{person}{Tinne Tuytelaars}.}
  \bibinfo{year}{2018}\natexlab{}.
\newblock \showarticletitle{{Memory Aware Synapses: Learning What (not) to
  Forget}}. In \bibinfo{booktitle}{\emph{{Proceedings of the European
  Conference on Computer Vision (ECCV)}}}. \bibinfo{pages}{139--154}.
\newblock


\bibitem[Ayub and Wagner(2020)]%
        {Ayub2020What}
\bibfield{author}{\bibinfo{person}{Ali Ayub} {and} \bibinfo{person}{Alan~R.
  Wagner}.} \bibinfo{year}{2020}\natexlab{}.
\newblock \showarticletitle{{What Am I Allowed to Do Here?: Online Learning of
  Context-Specific Norms by Pepper}}. In \bibinfo{booktitle}{\emph{Social
  Robotics}}, \bibfield{editor}{\bibinfo{person}{Alan~R. Wagner},
  \bibinfo{person}{David Feil-Seifer}, \bibinfo{person}{Kerstin~S. Haring},
  \bibinfo{person}{Silvia Rossi}, \bibinfo{person}{Thomas Williams},
  \bibinfo{person}{Hongsheng He}, {and} \bibinfo{person}{Shuzhi Sam~Ge}}
  (Eds.). \bibinfo{publisher}{Springer International Publishing},
  \bibinfo{pages}{220--231}.
\newblock


\bibitem[Benesty et~al\mbox{.}(2009)]%
        {benesty2009pearson}
\bibfield{author}{\bibinfo{person}{Jacob Benesty}, \bibinfo{person}{Jingdong
  Chen}, \bibinfo{person}{Yiteng Huang}, {and} \bibinfo{person}{Israel Cohen}.}
  \bibinfo{year}{2009}\natexlab{}.
\newblock \showarticletitle{Pearson correlation coefficient}.
\newblock In \bibinfo{booktitle}{\emph{Noise reduction in speech processing}}.
  \bibinfo{publisher}{Springer}, \bibinfo{pages}{37--40}.
\newblock


\bibitem[Chen and Liu(2018)]%
        {chen2018lifelong}
\bibfield{author}{\bibinfo{person}{Zhiyuan Chen} {and} \bibinfo{person}{Bing
  Liu}.} \bibinfo{year}{2018}\natexlab{}.
\newblock \bibinfo{booktitle}{\emph{Lifelong Machine Learning}}.
\newblock \bibinfo{publisher}{Springer International Publishing},
  \bibinfo{address}{Cham}.
\newblock


\bibitem[Churamani et~al\mbox{.}(2020)]%
        {Churamani2020CL4AR}
\bibfield{author}{\bibinfo{person}{Nikhil Churamani}, \bibinfo{person}{Sinan
  Kalkan}, {and} \bibinfo{person}{Hatice Gunes}.}
  \bibinfo{year}{2020}\natexlab{}.
\newblock \showarticletitle{{Continual Learning for Affective Robotics: Why,
  What and How?}}. In \bibinfo{booktitle}{\emph{{29th IEEE International
  Conference on Robot and Human Interactive Communication (RO-MAN)}}}. IEEE,
  \bibinfo{pages}{425--431}.
\newblock


\bibitem[Cramer et~al\mbox{.}(2010)]%
        {cramer2010effects}
\bibfield{author}{\bibinfo{person}{Henriette Cramer}, \bibinfo{person}{Jorrit
  Goddijn}, \bibinfo{person}{Bob Wielinga}, {and} \bibinfo{person}{Vanessa
  Evers}.} \bibinfo{year}{2010}\natexlab{}.
\newblock \showarticletitle{Effects of (in) accurate empathy and situational
  valence on attitudes towards robots}. In \bibinfo{booktitle}{\emph{2010 5th
  ACM/IEEE International Conference on Human-Robot Interaction (HRI)}}. IEEE,
  \bibinfo{pages}{141--142}.
\newblock


\bibitem[Francis et~al\mbox{.}(2023)]%
        {francis2023principles}
\bibfield{author}{\bibinfo{person}{Anthony Francis}, \bibinfo{person}{Claudia
  P{\'e}rez-d'Arpino}, \bibinfo{person}{Chengshu Li}, \bibinfo{person}{Fei
  Xia}, \bibinfo{person}{Alexandre Alahi}, \bibinfo{person}{Rachid Alami},
  \bibinfo{person}{Aniket Bera}, \bibinfo{person}{Abhijat Biswas},
  \bibinfo{person}{Joydeep Biswas}, \bibinfo{person}{Rohan Chandra},
  {et~al\mbox{.}}} \bibinfo{year}{2023}\natexlab{}.
\newblock \showarticletitle{Principles and guidelines for evaluating social
  robot navigation algorithms}.
\newblock \bibinfo{journal}{\emph{arXiv preprint arXiv:2306.16740}}
  (\bibinfo{year}{2023}).
\newblock


\bibitem[Gao et~al\mbox{.}(2019)]%
        {gao2019learning}
\bibfield{author}{\bibinfo{person}{Yuan Gao}, \bibinfo{person}{Fangkai Yang},
  \bibinfo{person}{Martin Frisk}, \bibinfo{person}{Daniel Hemandez},
  \bibinfo{person}{Christopher Peters}, {and} \bibinfo{person}{Ginevra
  Castellano}.} \bibinfo{year}{2019}\natexlab{}.
\newblock \showarticletitle{Learning socially appropriate robot approaching
  behavior toward groups using deep reinforcement learning}. In
  \bibinfo{booktitle}{\emph{2019 28th IEEE International Conference on Robot
  and Human Interactive Communication (RO-MAN)}}. IEEE, \bibinfo{pages}{1--8}.
\newblock


\bibitem[Guerdan and Gunes(2023)]%
        {Guerdan2023FCL}
\bibfield{author}{\bibinfo{person}{Luke Guerdan} {and} \bibinfo{person}{Hatice
  Gunes}.} \bibinfo{year}{2023}\natexlab{}.
\newblock \showarticletitle{Federated Continual Learning for Socially Aware
  Robotics}. In \bibinfo{booktitle}{\emph{2023 32nd IEEE International
  Conference on Robot and Human Interactive Communication (RO-MAN)}}.
  \bibinfo{pages}{1522--1529}.
\newblock


\bibitem[Hadsell et~al\mbox{.}(2020)]%
        {Hadsell2020CL}
\bibfield{author}{\bibinfo{person}{Raia Hadsell}, \bibinfo{person}{Dushyant
  Rao}, \bibinfo{person}{Andrei~A. Rusu}, {and} \bibinfo{person}{Razvan
  Pascanu}.} \bibinfo{year}{2020}\natexlab{}.
\newblock \showarticletitle{{Embracing Change: Continual Learning in Deep
  Neural Networks}}.
\newblock \bibinfo{journal}{\emph{{Trends in Cognitive Sciences}}}
  \bibinfo{volume}{24}, \bibinfo{number}{12} (\bibinfo{date}{Dec.}
  \bibinfo{year}{2020}), \bibinfo{pages}{1028--1040}.
\newblock


\bibitem[Hinton et~al\mbox{.}(2015)]%
        {Hinton2015Distillation}
\bibfield{author}{\bibinfo{person}{Geoffrey Hinton}, \bibinfo{person}{Oriol
  Vinyals}, {and} \bibinfo{person}{Jeffrey Dean}.}
  \bibinfo{year}{2015}\natexlab{}.
\newblock \showarticletitle{Distilling the Knowledge in a Neural Network}. In
  \bibinfo{booktitle}{\emph{NIPS Deep Learning and Representation Learning
  Workshop}}.
\newblock


\bibitem[Hsu et~al\mbox{.}(2018)]%
        {Hsu18_EvalCLNR}
\bibfield{author}{\bibinfo{person}{Yen-Chang Hsu}, \bibinfo{person}{Yen-Cheng
  Liu}, \bibinfo{person}{Anita Ramasamy}, {and} \bibinfo{person}{Zsolt Kira}.}
  \bibinfo{year}{2018}\natexlab{}.
\newblock \showarticletitle{{Re-evaluating Continual Learning Scenarios: A
  Categorization and Case for Strong Baselines}}. In
  \bibinfo{booktitle}{\emph{{NeurIPS Continual learning Workshop}}}.
\newblock


\bibitem[Imteaj et~al\mbox{.}(2022)]%
        {Imteaj2022Survey}
\bibfield{author}{\bibinfo{person}{Ahmed Imteaj}, \bibinfo{person}{Urmish
  Thakker}, \bibinfo{person}{Shiqiang Wang}, \bibinfo{person}{Jian Li}, {and}
  \bibinfo{person}{M.~Hadi Amini}.} \bibinfo{year}{2022}\natexlab{}.
\newblock \showarticletitle{A Survey on Federated Learning for
  Resource-Constrained IoT Devices}.
\newblock \bibinfo{journal}{\emph{IEEE Internet of Things Journal}}
  \bibinfo{volume}{9}, \bibinfo{number}{1} (\bibinfo{year}{2022}),
  \bibinfo{pages}{1--24}.
\newblock


\bibitem[Jiang et~al\mbox{.}(2020)]%
        {Jiang2020FedDistill}
\bibfield{author}{\bibinfo{person}{Donglin Jiang}, \bibinfo{person}{Chen Shan},
  {and} \bibinfo{person}{Zhihui Zhang}.} \bibinfo{year}{2020}\natexlab{}.
\newblock \showarticletitle{Federated Learning Algorithm Based on Knowledge
  Distillation}.
\newblock \bibinfo{journal}{\emph{2020 International Conference on Artificial
  Intelligence and Computer Engineering (ICAICE)}}, \bibinfo{pages}{163--167}.
\newblock
\showISBNx{978-1-7281-9146-1}


\bibitem[Kirkpatrick et~al\mbox{.}(2017)]%
        {kirkpatrick2017overcomingEWC}
\bibfield{author}{\bibinfo{person}{James Kirkpatrick}, \bibinfo{person}{Razvan
  Pascanu}, \bibinfo{person}{Neil Rabinowitz}, \bibinfo{person}{Joel Veness},
  \bibinfo{person}{Guillaume Desjardins}, \bibinfo{person}{Andrei~A Rusu},
  \bibinfo{person}{Kieran Milan}, \bibinfo{person}{John Quan},
  \bibinfo{person}{Tiago Ramalho}, \bibinfo{person}{Agnieszka
  Grabska-Barwinska}, \bibinfo{person}{Demis Hassabis},
  \bibinfo{person}{Claudia Clopath}, \bibinfo{person}{Dharshan Kumaran}, {and}
  \bibinfo{person}{Raia Hadsell}.} \bibinfo{year}{2017}\natexlab{}.
\newblock \showarticletitle{{Overcoming catastrophic forgetting in neural
  networks}}.
\newblock \bibinfo{journal}{\emph{{Proceedings of the National Academy of
  Sciences}}} \bibinfo{volume}{114}, \bibinfo{number}{13}
  (\bibinfo{year}{2017}), \bibinfo{pages}{3521--3526}.
\newblock
\showISSN{0027-8424}


\bibitem[Lesort et~al\mbox{.}(2020)]%
        {LESORT2020CL4R}
\bibfield{author}{\bibinfo{person}{Timothée Lesort}, \bibinfo{person}{Vincenzo
  Lomonaco}, \bibinfo{person}{Andrei Stoian}, \bibinfo{person}{Davide Maltoni},
  \bibinfo{person}{David Filliat}, {and} \bibinfo{person}{Natalia
  Díaz-Rodríguez}.} \bibinfo{year}{2020}\natexlab{}.
\newblock \showarticletitle{{Continual learning for robotics: Definition,
  framework, learning strategies, opportunities and challenges}}.
\newblock \bibinfo{journal}{\emph{{Information Fusion}}}  \bibinfo{volume}{58}
  (\bibinfo{year}{2020}), \bibinfo{pages}{52--68}.
\newblock
\showISSN{1566-2535}


\bibitem[Li et~al\mbox{.}(2020b)]%
        {Li2020FedProx}
\bibfield{author}{\bibinfo{person}{Tian Li}, \bibinfo{person}{Anit~Kumar Sahu},
  \bibinfo{person}{Manzil Zaheer}, \bibinfo{person}{Maziar Sanjabi},
  \bibinfo{person}{Ameet Talwalkar}, {and} \bibinfo{person}{Virginia Smith}.}
  \bibinfo{year}{2020}\natexlab{b}.
\newblock \showarticletitle{Federated Optimization in Heterogeneous Networks}.
  In \bibinfo{booktitle}{\emph{Proceedings of Machine Learning and Systems}},
  \bibfield{editor}{\bibinfo{person}{I.~Dhillon},
  \bibinfo{person}{D.~Papailiopoulos}, {and} \bibinfo{person}{V.~Sze}} (Eds.),
  Vol.~\bibinfo{volume}{2}. \bibinfo{pages}{429--450}.
\newblock


\bibitem[Li et~al\mbox{.}(2020a)]%
        {Li2020FedAvg}
\bibfield{author}{\bibinfo{person}{Xiang Li}, \bibinfo{person}{Kaixuan Huang},
  \bibinfo{person}{Wenhao Yang}, \bibinfo{person}{Shusen Wang}, {and}
  \bibinfo{person}{Zhihua Zhang}.} \bibinfo{year}{2020}\natexlab{a}.
\newblock \showarticletitle{On the Convergence of FedAvg on Non-IID Data}. In
  \bibinfo{booktitle}{\emph{International Conference on Learning
  Representations}}.
\newblock


\bibitem[Li et~al\mbox{.}(2021)]%
        {Li2021FedBN}
\bibfield{author}{\bibinfo{person}{Xiaoxiao Li}, \bibinfo{person}{Meirui
  JIANG}, \bibinfo{person}{Xiaofei Zhang}, \bibinfo{person}{Michael Kamp},
  {and} \bibinfo{person}{Qi Dou}.} \bibinfo{year}{2021}\natexlab{}.
\newblock \showarticletitle{Fed{BN}: Federated Learning on Non-{IID} Features
  via Local Batch Normalization}. In \bibinfo{booktitle}{\emph{International
  Conference on Learning Representations}}.
\newblock


\bibitem[Liu et~al\mbox{.}(2020)]%
        {Liu2020Geneative}
\bibfield{author}{\bibinfo{person}{Xialei Liu}, \bibinfo{person}{Chenshen Wu},
  \bibinfo{person}{Mikel Menta}, \bibinfo{person}{Luis Herranz},
  \bibinfo{person}{Bogdan Raducanu}, \bibinfo{person}{Andrew~D. Bagdanov},
  \bibinfo{person}{Shangling Jui}, {and} \bibinfo{person}{Joost van~de
  Weijer}.} \bibinfo{year}{2020}\natexlab{}.
\newblock \showarticletitle{Generative Feature Replay For Class-Incremental
  Learning}. In \bibinfo{booktitle}{\emph{{IEEE}/{CVF} Conference on Computer
  Vision and Pattern Recognition Workshops ({CVPRW})}}.
  \bibinfo{publisher}{{IEEE}}.
\newblock


\bibitem[Manso et~al\mbox{.}(2020)]%
        {Manso2020SocNav1}
\bibfield{author}{\bibinfo{person}{Luis~J. Manso}, \bibinfo{person}{Pedro
  Nuñez}, \bibinfo{person}{Luis~V. Calderita}, \bibinfo{person}{Diego~R.
  Faria}, {and} \bibinfo{person}{Pilar Bachiller}.}
  \bibinfo{year}{2020}\natexlab{}.
\newblock \showarticletitle{SocNav1: A Dataset to Benchmark and Learn Social
  Navigation Conventions}.
\newblock \bibinfo{journal}{\emph{Data}} \bibinfo{volume}{5},
  \bibinfo{number}{1} (\bibinfo{year}{2020}).
\newblock
\showISSN{2306-5729}


\bibitem[McMahan et~al\mbox{.}(2017)]%
        {McMahan2017FL}
\bibfield{author}{\bibinfo{person}{Brendan McMahan}, \bibinfo{person}{Eider
  Moore}, \bibinfo{person}{Daniel Ramage}, \bibinfo{person}{Seth Hampson},
  {and} \bibinfo{person}{Blaise Aguera~y Arcas}.}
  \bibinfo{year}{2017}\natexlab{}.
\newblock \showarticletitle{{Communication-Efficient Learning of Deep Networks
  from Decentralized Data}}. In \bibinfo{booktitle}{\emph{Proceedings of the
  20th International Conference on Artificial Intelligence and Statistics}}
  \emph{(\bibinfo{series}{Proceedings of Machine Learning Research},
  Vol.~\bibinfo{volume}{54})}, \bibfield{editor}{\bibinfo{person}{Aarti Singh}
  {and} \bibinfo{person}{Jerry Zhu}} (Eds.). \bibinfo{publisher}{PMLR},
  \bibinfo{pages}{1273--1282}.
\newblock


\bibitem[McQuillin et~al\mbox{.}(2022)]%
        {McQuillin2022RoboWaiter}
\bibfield{author}{\bibinfo{person}{Emily McQuillin}, \bibinfo{person}{Nikhil
  Churamani}, {and} \bibinfo{person}{Hatice Gunes}.}
  \bibinfo{year}{2022}\natexlab{}.
\newblock \showarticletitle{{Learning Socially Appropriate Robo-Waiter
  Behaviours through Real-Time User Feedback}}. In
  \bibinfo{booktitle}{\emph{{Proceedings of the 2022 ACM/IEEE International
  Conference on Human-Robot Interaction (HRI)}}} (Sapporo, Hokkaido, Japan)
  \emph{(\bibinfo{series}{HRI '22})}. \bibinfo{publisher}{IEEE Press},
  \bibinfo{pages}{541–550}.
\newblock


\bibitem[Parisi et~al\mbox{.}(2019)]%
        {Parisi2018b}
\bibfield{author}{\bibinfo{person}{German~I. Parisi}, \bibinfo{person}{Ronald
  Kemker}, \bibinfo{person}{Jose~L. Part}, \bibinfo{person}{Christopher Kanan},
  {and} \bibinfo{person}{Stefan Wermter}.} \bibinfo{year}{2019}\natexlab{}.
\newblock \showarticletitle{{Continual Lifelong Learning with Neural Networks:
  A review}}.
\newblock \bibinfo{journal}{\emph{Neural Networks}}  \bibinfo{volume}{113}
  (\bibinfo{year}{2019}), \bibinfo{pages}{54--71}.
\newblock


\bibitem[Parmar et~al\mbox{.}(2023)]%
        {Parmar2023OWL}
\bibfield{author}{\bibinfo{person}{Jitendra Parmar}, \bibinfo{person}{Satyendra
  Chouhan}, \bibinfo{person}{Vaskar Raychoudhury}, {and}
  \bibinfo{person}{Santosh Rathore}.} \bibinfo{year}{2023}\natexlab{}.
\newblock \showarticletitle{Open-World Machine Learning: Applications,
  Challenges, and Opportunities}.
\newblock \bibinfo{journal}{\emph{ACM Comput. Surv.}} \bibinfo{volume}{55},
  \bibinfo{number}{10}, Article \bibinfo{articleno}{205} (\bibinfo{date}{feb}
  \bibinfo{year}{2023}), \bibinfo{numpages}{37}~pages.
\newblock
\showISSN{0360-0300}


\bibitem[Reddi et~al\mbox{.}(2021)]%
        {Reddi2021FedOpt}
\bibfield{author}{\bibinfo{person}{Sashank~J. Reddi}, \bibinfo{person}{Zachary
  Charles}, \bibinfo{person}{Manzil Zaheer}, \bibinfo{person}{Zachary Garrett},
  \bibinfo{person}{Keith Rush}, \bibinfo{person}{Jakub Kone{\v{c}}n{\'y}},
  \bibinfo{person}{Sanjiv Kumar}, {and} \bibinfo{person}{Hugh~Brendan
  McMahan}.} \bibinfo{year}{2021}\natexlab{}.
\newblock \showarticletitle{Adaptive Federated Optimization}. In
  \bibinfo{booktitle}{\emph{International Conference on Learning
  Representations}}.
\newblock


\bibitem[Schwarz et~al\mbox{.}(2018)]%
        {schwarz2018progressEWCOnline}
\bibfield{author}{\bibinfo{person}{Jonathan Schwarz}, \bibinfo{person}{Wojciech
  Czarnecki}, \bibinfo{person}{Jelena Luketina}, \bibinfo{person}{Agnieszka
  Grabska-Barwinska}, \bibinfo{person}{Yee~Whye Teh}, \bibinfo{person}{Razvan
  Pascanu}, {and} \bibinfo{person}{Raia Hadsell}.}
  \bibinfo{year}{2018}\natexlab{}.
\newblock \showarticletitle{{Progress \& Compress: A scalable framework for
  continual learning}}. In \bibinfo{booktitle}{\emph{{Proceedings of the 35th
  International Conference on Machine Learning}}}
  \emph{(\bibinfo{series}{Proceedings of Machine Learning Research},
  Vol.~\bibinfo{volume}{80})}, \bibfield{editor}{\bibinfo{person}{Jennifer Dy}
  {and} \bibinfo{person}{Andreas Krause}} (Eds.). \bibinfo{publisher}{PMLR},
  \bibinfo{address}{Stockholmsmässan, Stockholm Sweden},
  \bibinfo{pages}{4528--4537}.
\newblock


\bibitem[Stoychev et~al\mbox{.}(2023)]%
        {Stoychev2023LGR}
\bibfield{author}{\bibinfo{person}{Samuil Stoychev}, \bibinfo{person}{Nikhil
  Churamani}, {and} \bibinfo{person}{Hatice Gunes}.}
  \bibinfo{year}{2023}\natexlab{}.
\newblock \showarticletitle{Latent Generative Replay for Resource-Efficient
  Continual Learning of Facial Expressions}. In \bibinfo{booktitle}{\emph{17th
  IEEE International Conference on Automatic Face and Gesture Recognition
  (FG)}}. \bibinfo{pages}{1--8}.
\newblock


\bibitem[Tjomsland et~al\mbox{.}(2022)]%
        {tjomsland2022mind}
\bibfield{author}{\bibinfo{person}{Jonas Tjomsland}, \bibinfo{person}{Sinan
  Kalkan}, {and} \bibinfo{person}{Hatice Gunes}.}
  \bibinfo{year}{2022}\natexlab{}.
\newblock \showarticletitle{{Mind Your Manners! A Dataset and a Continual
  Learning Approach for Assessing Social Appropriateness of Robot Actions}}.
\newblock \bibinfo{journal}{\emph{Frontiers in Robotics and AI}}
  \bibinfo{volume}{9} (\bibinfo{year}{2022}).
\newblock
\showISSN{2296-9144}


\bibitem[Yoon et~al\mbox{.}(2021)]%
        {Yoon2021FCL}
\bibfield{author}{\bibinfo{person}{Jaehong Yoon}, \bibinfo{person}{Wonyong
  Jeong}, \bibinfo{person}{Giwoong Lee}, \bibinfo{person}{Eunho Yang}, {and}
  \bibinfo{person}{Sung~Ju Hwang}.} \bibinfo{year}{2021}\natexlab{}.
\newblock \showarticletitle{Federated Continual Learning with Weighted
  Inter-client Transfer}. In \bibinfo{booktitle}{\emph{Proceedings of the 38th
  International Conference on Machine Learning}}
  \emph{(\bibinfo{series}{Proceedings of Machine Learning Research},
  Vol.~\bibinfo{volume}{139})}, \bibfield{editor}{\bibinfo{person}{Marina
  Meila} {and} \bibinfo{person}{Tong Zhang}} (Eds.). \bibinfo{publisher}{PMLR},
  \bibinfo{pages}{12073--12086}.
\newblock


\bibitem[Zenke et~al\mbox{.}(2017)]%
        {zenke2017continualSI}
\bibfield{author}{\bibinfo{person}{Friedemann Zenke}, \bibinfo{person}{Ben
  Poole}, {and} \bibinfo{person}{Surya Ganguli}.}
  \bibinfo{year}{2017}\natexlab{}.
\newblock \showarticletitle{{Continual Learning Through Synaptic
  Intelligence}}, In \bibinfo{booktitle}{{Proceedings of the 34th International
  Conference on Machine Learning - Volume 70}} (Sydney, NSW, Australia).
\newblock \bibinfo{journal}{\emph{{Proceedings of machine learning research}}}
  \bibinfo{volume}{70}, \bibinfo{pages}{3987--3995}.
\newblock


\bibitem[Zhang et~al\mbox{.}(2021)]%
        {Zhang2021Survey}
\bibfield{author}{\bibinfo{person}{Chen Zhang}, \bibinfo{person}{Yu Xie},
  \bibinfo{person}{Hang Bai}, \bibinfo{person}{Bin Yu},
  \bibinfo{person}{Weihong Li}, {and} \bibinfo{person}{Yuan Gao}.}
  \bibinfo{year}{2021}\natexlab{}.
\newblock \showarticletitle{A survey on federated learning}.
\newblock \bibinfo{journal}{\emph{Knowledge-Based Systems}}
  \bibinfo{volume}{216} (\bibinfo{year}{2021}), \bibinfo{pages}{106775}.
\newblock
\showISSN{0950-7051}


\end{thebibliography}

\end{document}